\documentclass[10pt,twocolumn,letterpaper]{article}

\usepackage{cvpr}              

\newcommand{\red}[1]{{\color{red}#1}}

\definecolor{cvprblue}{rgb}{0.21,0.49,0.74}
\usepackage[pagebackref,breaklinks,colorlinks,allcolors=cvprblue]{hyperref}

\usepackage{bm}
\usepackage[accsupp]{axessibility}

\def\paperID{*****} 
\def\confName{CVPR}
\def\confYear{2026}

\title{End-to-End Shared Attention Estimation \\ via Group Detection with Feedback Refinement}

\author{Chihiro Nakatani$^{1}$\quad Norimichi Ukita$^{1}$\quad Jean-Marc Odobez$^{2,3}$\\
$^{1}$ Toyota Technological Institute, Japan \quad $^{2}$ Idiap Research Institute, Switzerland \\ \quad $^{3}$ \'Ecole Polytechnique F\'ed\'erale de Lausanne, Switzerland}

\begin{document}
\maketitle
\begin{abstract}
This paper proposes an end-to-end shared attention estimation method via group detection. Most previous methods estimate shared attention (SA) without detecting the actual group of people focusing on it, or assume that there is a single SA point in a given image. These issues limit the applicability of SA detection in practice and impact performance. To address them, we propose to simultaneously achieve group detection and shared attention estimation using a two step process: (i) the generation of SA heatmaps relying on individual gaze attention heatmaps and group membership scalars estimated in a group inference; (ii) a refinement of the initial group memberships allowing to account for the initial SA heatmaps, and the final prediction of the SA heatmap. Experiments demonstrate that our method outperforms other methods in group detection and shared attention estimation. Additional analyses validate the effectiveness of the proposed components. Code: \url{https://github.com/chihina/sagd-CVPRW2026}.
\end{abstract}
\section{Introduction}
\label{sec:intro}

Human attention analysis is an essential topic for understanding people's behaviors and social interactions. 
The basic unit related to human attention is defined by the gaze target of people, defined as the points at which people are looking. 
This topic
has been 
investigated
in~\cite{DBLP:conf/nips/RecasensKVT15,DBLP:conf/iccv/RecasensVK017,DBLP:conf/eccv/ChongRWZRR18,DBLP:journals/tcsv/ChenXZLLZK22,DBLP:conf/cvpr/ChongWRR20,DBLP:conf/cvpr/FangTSS00Z21,DBLP:conf/cvpr/TuMDGZS22,DBLP:conf/cvpr/HuYZYZL23,DBLP:conf/iccv/TafascaGO23,DBLP:conf/cvpr/GuptaTO22,DBLP:conf/cvpr/TafascaGO24,DBLP:conf/nips/GuptaTFVO24,DBLP:conf/cvpr/RyanB0BHR25,DBLP:conf/cvpr/HoranyiZCLC23,DBLP:conf/eccv/YangL24,DBLP:conf/aaai/YangYL24,DBLP:conf/nips/TafascaGO24}.
These methods estimate each person's 2D gaze target by relying on the target person's cues (e.g., a head image) and the scene context captured across the entire image.

Beyond individual attention, identifying key social gaze cues like whether two persons are in eye contact or a group of people have a shared attention towards an object is important for understanding people's interactions or group behavior in various applications (e.g., in HRI, a robot naturally intervening in a group conversation~\cite{DBLP:conf/iros/SaranMSTN18}; 
in the health domain, analyzing gaze patterns for assessing autism spectrum disorders~\cite{DBLP:conf/iccv/ChenZ19,DBLP:conf/iccv/JiangZ17,DBLP:conf/icmi/TafascaGKGMPSO23}).

\begin{figure}[t]
    \centering
    \includegraphics[width=\columnwidth]{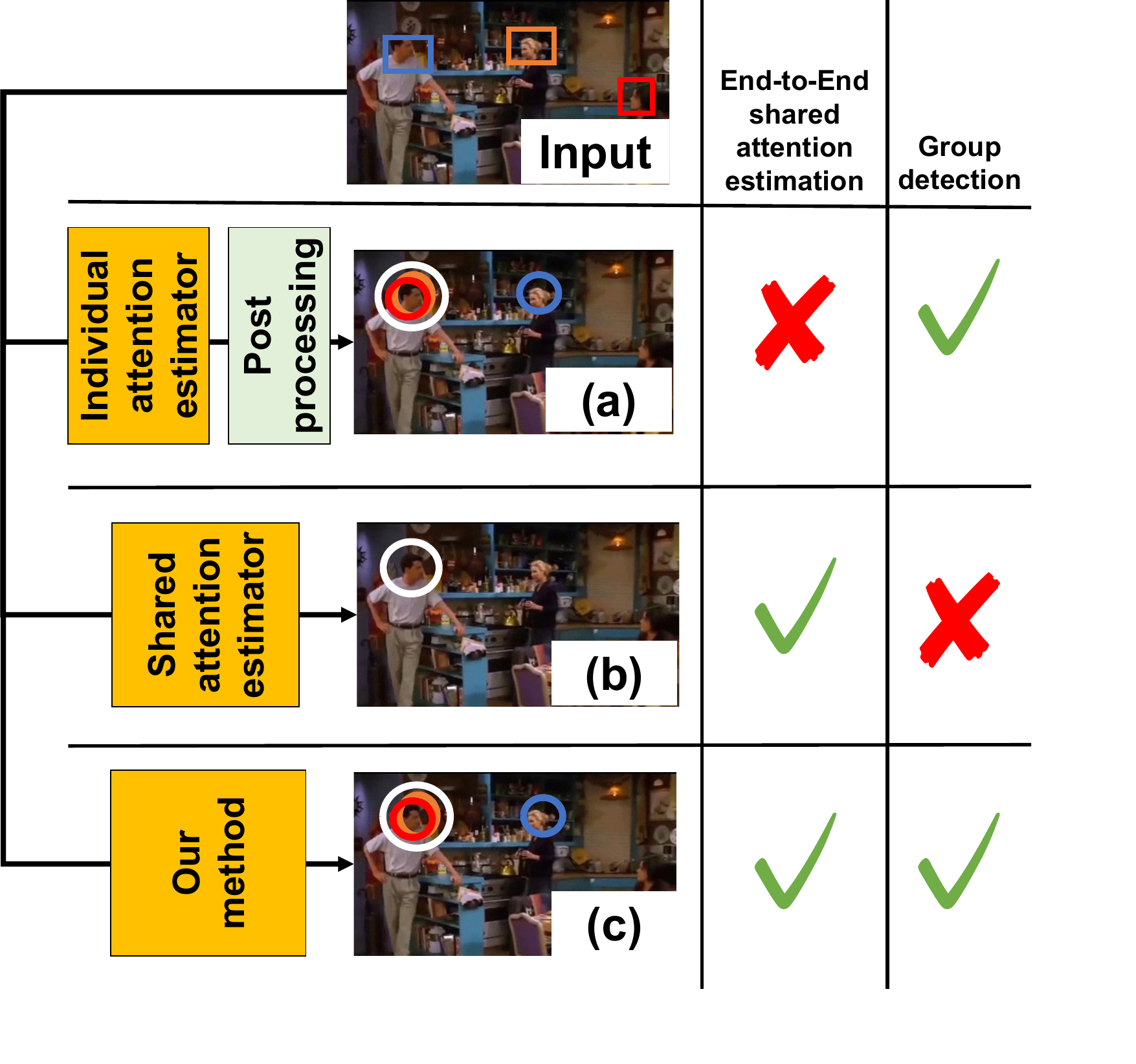}
    \caption{Difference between previous and our shared attention estimation methods. 
    (a) Shared attention estimation using simple integration of the individual attention of all people in the scene as post-processing. 
    (b) Direct shared attention estimation without group detection.
    (c) Our shared attention estimation via group detection, in which shared attention is estimated by integration of individual attention based on detected groups.
    }
\end{figure}

The growth of interest in deriving attention 
cues has motivated researchers to further analyze complex shared attention situations involving multiple people.  
One approach towards this goal is to integrate (e.g., average) the single-person attentions heatmap of each person in a scene, as used in~\cite{DBLP:conf/cvpr/ChongWRR20}.
However, such a simple integration fails to capture some form of group context in the scene.
Directly estimating shared attention from multi-person cues is also proposed in~\cite{DBLP:conf/cvpr/FanCWWZ18,DBLP:journals/tcsv/ZhuangNXYZLG20,DBLP:conf/wacv/SumerGTK20,DBLP:conf/iccv/NakataniKU23}.
Although the estimation is better than the simple approach in terms of group context, these methods are limited to estimating shared attention points without detecting which group members are participating in it.
To simultaneously incorporate the use of group context and the detection of group members, Anshul~\textit{et al}.~\cite{DBLP:conf/nips/GuptaTFVO24} proposed a pairwise approach. 
This method decomposes shared attention estimation into pairwise estimations. Instead of directly estimating the people participating in shared attention, this method estimates whether pairs of  people share their attention, an inference made across all pair combinations in a scene. 
Although this approach can exploit group context, it may lead to inconsistent results: Person A may share attention with B, and B with C, but A and C might be inferred as not sharing attention. 

Based on the discussion above, we can summarize that all previous approaches lack the ability to properly exploit group context for shared attention estimation in a consistent manner, and to consolidate individual attention into common shared attention targets. 
These limitations motivate us to develop a new method that estimates shared attention via group detection to model group context.
In contrast to the pairwise approaches, our method proposes a novel method that detects whole groups of people sharing attention through global optimization.

To this end, we propose to jointly infer shared-attention heatmaps along with shared-attention membership identification.  
More specifically, single-person attention heatmaps are first estimated based on each person's cues and the scene information, similarly to previous single-person attention estimation methods.
Then, our method detects groups of people sharing attention through the inference of group membership scalars, which represent the likelihood of individuals focusing on the same target.
These estimated memberships are subsequently used to infer shared attention heatmaps for each group simply by computing a membership-weighted sum of single-person attention heatmaps.
This membership-based shared attention heatmapping enables an end-to-end shared attention estimation via group detection.

Our contributions are summarized as follows:
\begin{itemize}
\item {\bf End-to-End shared attention estimation via group detection:}
Unlike previous methods that separately address shared attention estimation and group detection, this paper proposes estimating the shared attention via group detection, enabling the full capture of the group context.

\item {\bf Group membership approach for shared attention heatmap estimation:} 
Specifically, we generate each group's shared attention heatmap by weighting and aggregating individual attention maps based on their estimated shared attention group memberships.

\item {\bf Shared attention refinement:} 
In the forward pipeline described above, initial group detection memberships can not rely on the shared attention heatmap knowledge. This motivates us to introduce a refining group membership detection step and a final shared attention heatmap estimation step. 

\end{itemize}
\section{Related Work}
\label{sec:rel_work}

\subsection{Individual Attention Estimation}
\label{subsec:single_att}

Individual attention estimation predicts a person's attention from individual appearance cues and global scene context.

Early work~\cite{DBLP:conf/nips/RecasensKVT15,DBLP:conf/iccv/RecasensVK017,DBLP:conf/eccv/ChongRWZRR18,DBLP:conf/cvpr/ChongWRR20} estimates the attention of a target person from their head and the whole image.
In~\cite{DBLP:conf/cvpr/FangTSS00Z21}, a depth image estimated from a whole image is used as an auxiliary cue.
Tafasca~\textit{et al}.~\cite{DBLP:conf/iccv/TafascaGO23} utilize a scene point cloud to obtain 3DFoV information for estimation.
Unlike these methods which leverage various modalities, Ryan~\textit{et al}.~\cite{DBLP:conf/cvpr/RyanB0BHR25} use DINO~\cite{DBLP:journals/tmlr/OquabDMVSKFHMEA24} as a powerful backbone.
While these methods focus only on a single person's attention, the features of all people in a scene are used to jointly infer the attentions of all people in a scene, implicitly modeling interactions in~\cite{DBLP:conf/cvpr/TafascaGO24}.
In addition to individual attention, Gupta~\textit{et al}.~\cite{DBLP:conf/nips/GuptaTFVO24} further predict social labels in a pairwise manner (e.g., whether two people share their attention or not) to better capture interactions.

While these individual attention estimations can be extended to shared attention estimation, it may lead to inconsistent results as mentioned in Sec.~\ref{sec:intro}.

\begin{figure*}[t]
\begin{center}
\includegraphics[width=\textwidth]{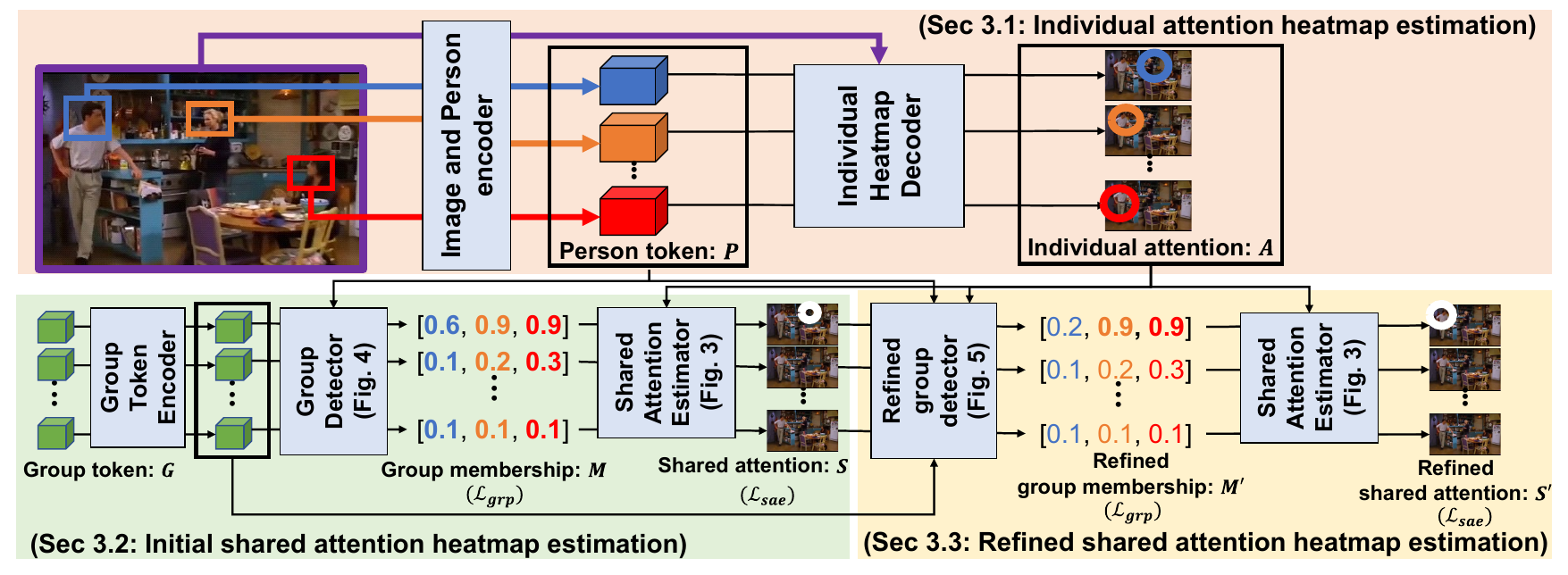}\\
\end{center}
\vspace{-5mm}
\caption{Overview of our network. First, the individual attention heatmaps $A$ are estimated for each person. They are then exploited to derive group memberships per each group token $\bm{M}$ and integrated to infer the shared attention heatmap associated with each group token $\bm{S}$. 
Both the group memberships and shared attention heatmaps 
are further refined in a second step. 
}
\label{fig:overview}
\end{figure*}

\subsection{Shared Attention Estimation}
\label{subsec:shared_att}

Beyond extending single-person attention estimation to shared attention estimation, various approaches~\cite{DBLP:conf/cvpr/FanCWWZ18,DBLP:journals/tcsv/ZhuangNXYZLG20,DBLP:conf/iccv/NakataniKU23,DBLP:conf/wacv/SumerGTK20} focus on directly estimating shared attention.
In~\cite{DBLP:conf/cvpr/FanCWWZ18}, cone-shaped gaze heatmaps generated for each person based on their gaze directions are fed into a CNN network to estimate the shared attention heatmap.
In~\cite{DBLP:journals/tcsv/ZhuangNXYZLG20}, the coordinates of shared attention are also estimated from the cone-shaped gaze heatmaps as regression in addition to the heatmapping of shared attention.
Sumer~\textit{et al.}~\cite{DBLP:conf/wacv/SumerGTK20} utilize saliency maps, which can be obtained from each image, to augment shared attention estimation.
Different from these methods, Nakatani~\textit{et al}.~\cite{DBLP:conf/iccv/NakataniKU23} employs multiple attributes (e.g., locations, gaze directions, and actions) of each person as cues to estimate the shared attention heatmaps.
 
While all of these methods focus solely on estimating shared attention, our method simultaneously estimates shared attention heatmaps and specific group members involved, thereby capturing group context more effectively.

\subsection{Group Detection}
\label{subsec:group_det}

Since group detection is a fundamental task for understanding high-level context within groups, various group detection methods have been proposed in~\cite{DBLP:conf/eccv/LiHYQFW22,DBLP:conf/iccv/YokoyamaKU25,DBLP:conf/eccv/KimSCK24}.
In~\cite{DBLP:conf/eccv/LiHYQFW22}, self-supervised learning is applied to learn multiple people's behavior as pre-training for subsequent group detection in a supervised manner. 
In~\cite{DBLP:conf/iccv/YokoyamaKU25}, CLIP is fine-tuned for group detection to capture scene context from a whole image.
Kim~\textit{et al}.~\cite{DBLP:conf/eccv/KimSCK24} have proposed joint learning of group detection and activity recognition on each detected group. 

While Kim~\textit{et al}.~\cite{DBLP:conf/eccv/KimSCK24} is similar to our method in that it incorporates a group detection, the downstream task (i.e., activity recognition in~\cite{DBLP:conf/eccv/KimSCK24}) is different. 
Furthermore, our method refines group detection based on initial shared attention estimation, whereas Kim~\textit{et al}. estimate the group and its activity in a one-way pipeline without considering interaction between group detection and activity recognition.
\section{Proposed Method}
\label{sec:prop_method}

\vspace*{-1mm}

An overview of our approach is provided in Fig.~\ref{fig:overview}.
In the individual attention estimation step (Sec.~\ref{subsec:prop_single_att}), the gaze tokens of each person are encoded from their head crop and bounding box coordinates, and used as input to a visual backbone along with the image token to produce the person tokens $P$ used to decode the individual attention heatmaps, as proposed in~\cite{DBLP:conf/nips/GuptaTFVO24}. 
In the initial SA estimation step (Sec.~\ref{subsec:prop_init_sae}), the individual attention heatmaps are integrated using group membership to estimate initial SA heatmaps. 
Group detection is achieved thanks to the use of learnable group tokens, each representing a potential group. 
Group membership coefficients for all people in the scene are associated with each of these tokens (initially, by computing the similarity between these tokens and the person tokens) and used to derive a SA heatmap for this group.
In the refined SA estimation step (Sec.~\ref{subsec:prop_refine_sae}), the estimated initial SA heatmaps are further exploited to refine the group memberships and recompute a final SA heatmap.

\begin{figure}[t]
\begin{center}
\includegraphics[width=\columnwidth]{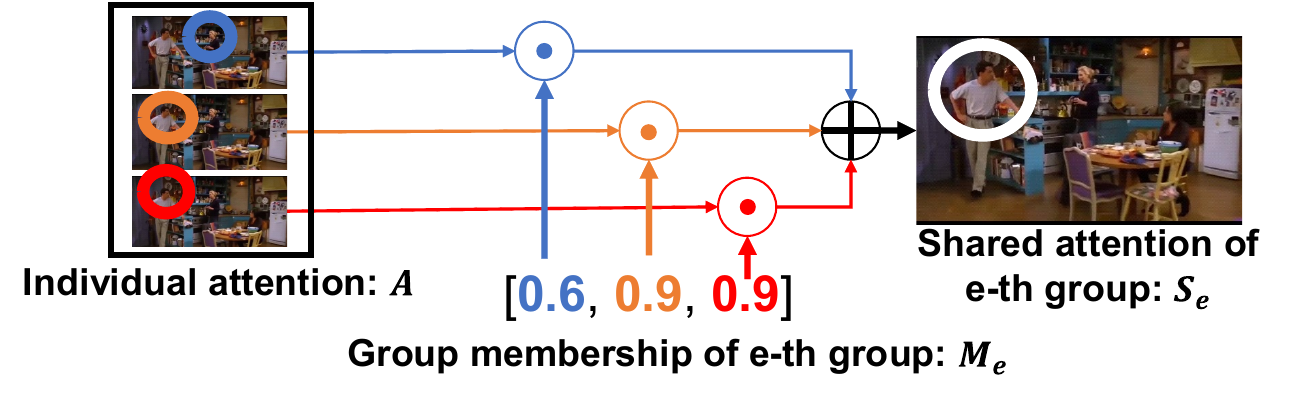}\\
\end{center}
\vspace*{-4mm}
\caption{Membership-based shared attention heatmap estimation. Individual attentions heatmaps are integrated by weighting them using the group membership obtained in the group detection step.}
\label{fig:sae_merge}
\vspace*{-2mm}
\end{figure}

\subsection{Individual attention heatmap estimation}
\label{subsec:prop_single_att}

\vspace*{-1mm}

\paragraph{Key idea.}
Our method estimates an individual attention heatmap for each person in a scene and integrates them to compute a shared attention heatmap (Sec.~\ref{subsec:prop_init_sae}).
The intermediate token produced in this step will be used in later stages to infer SA group memberships. 

\vspace*{-4mm}

\paragraph{Details.}
For each person, the head bounding box coordinates and the cropped head image are first encoded as a person gaze token.
These person tokens are then processed by transformers along with image tokens using person-person and person-scene self and cross-attention interaction module as proposed in~\cite{DBLP:conf/nips/GuptaTFVO24}. 
This results in the person tokens $\bm{P} \in \mathbb{R}^{N \times D}$ where $N$ denotes the number of people in a scene, and $D$ is the dimension of each person token.

In a second step, for the $n$-th person, $\bm{P}_{n}$ and multiresolution features from the whole image computed during the above encoding stage are fed into the individual heatmap decoder to obtain the individual attention heatmap denoted by $\bm{A}_{n} \in \mathbb{R}^{H \times W}$. See ~\cite{DBLP:conf/nips/GuptaTFVO24} for more details. 

\begin{figure}[t]
\begin{center}
\includegraphics[width=\columnwidth]{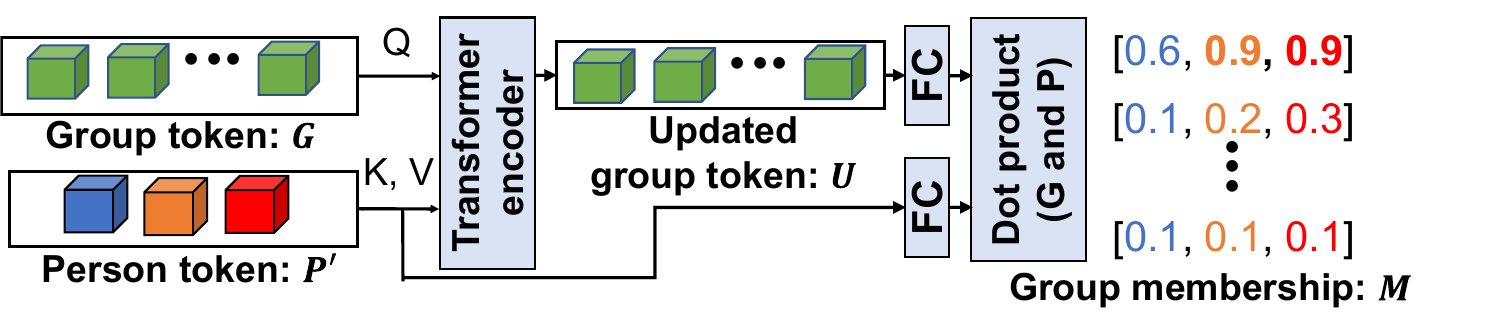}\\
\end{center}
\vspace{-2mm}
\caption{
Overview of group detection. 
The dot product between the $e$-th updated group token and the $n$-th person token is computed as a group membership coefficient (i.e., $\bm{M}_{e,n}$).
}
\label{fig:member_estimation}
\end{figure}

\begin{figure}[t]
\begin{center}
\includegraphics[width=\columnwidth]{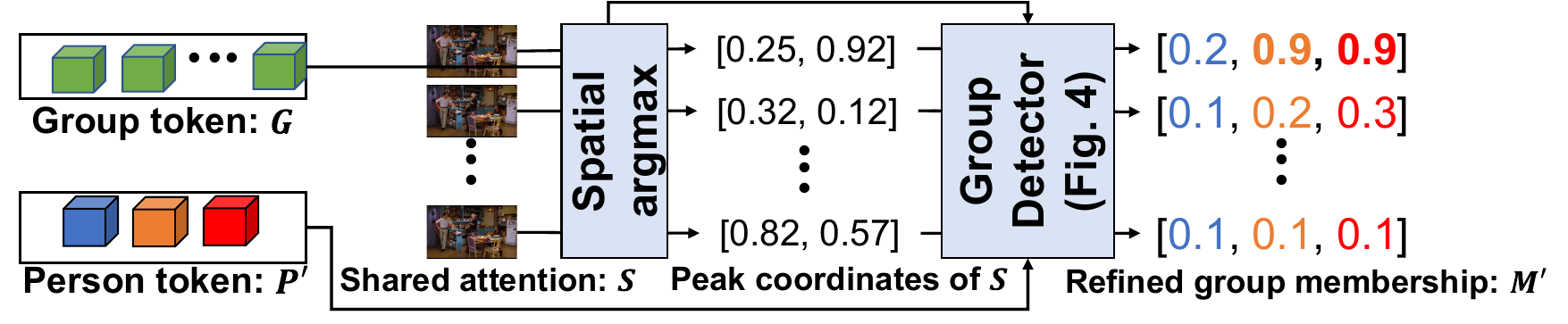}\\
\end{center}
\vspace{-5mm}
\caption{Overview of refined group detection in which spatial argmax is applied to the initial SA heatmaps $\bm{S}$ for refinement. 
}
\label{fig:sae_refinement}
\end{figure}

\subsection{Initial Shared Attention Heatmap Estimation}
\label{subsec:prop_init_sae}

\paragraph{Key idea.}
Our method integrates estimated individual attention heatmaps via group detection as shown in Fig.~\ref{fig:sae_merge}.
The group-detection-based integration enables simultaneous estimation of the shared attention heatmap and the people focusing on it, 
a step that is often completely overlooked in previous 
shared attention estimation methods~\cite{DBLP:conf/cvpr/FanCWWZ18,DBLP:journals/tcsv/ZhuangNXYZLG20}.

\vspace{-2mm}

\paragraph{Details.}
Our method initializes learnable group tokens $\bm{G} \in \mathbb{R}^{E \times D}$ as with DETR~\cite{DBLP:conf/cvpr/TuMDGZS22}, 
where $E$ denotes the maximum number of groups sharing attention in the image. 
The group tokens $\bm{G}$ are processed by a transformer-based encoder to capture relationships within potential groups. Subsequently, the encoded $\bm{G}$ is processed by the transformer-based group detector to model interactions with 
$\bm{P}$ resulting from individual heatmap encoding.
More specifically, the token of each person $P_n$ is further augmented to contain the more precise attention inferred for this person by directly concatenating the $(x,y)$ coordinates of the peak attention of the person in $\bm{A}$, resulting in 
 $\bm{P_n'} \in \mathbb{R}^{D+2}$.
In this transformer, $\bm{G}$ plays the role of  
the query, while $\bm{P'}$ are the keys and values as shown in Fig.~\ref{fig:member_estimation}.
Intuitively, through this cross-attention layer, the group tokens $\bm{G}$ are updated to capture both the individuals who might share the same attention and update the potential shared attention location information.
We denote by  $\bm{U}$ the updated group tokens. 

In a second step, the membership $\bm{M} \in \mathbb{R}^{E \times N}$, where $\bm{M}_{e,n}$ indicates how likely the $n$-th person belongs to the $e$-th group, is computed as follows:
\begin{align}
    \bm{M}_{e,n} = \mathrm{Sigmoid} (\mathrm{g}(U_{e}) \cdot \mathrm{f}(P'_{n}))
    \label{eq:groupness}
\end{align}
where $\mathrm{g}$ and $\mathrm{f}$ denote FC layers to embed $\bm{U}$ and $\bm{P'}$ in the same semantic space allowing to identify the memberships $\bm{M}_{e,n}$.
Sigmoids allow a person to belong to multiple groups, capturing scenarios where individuals simultaneously join multiple groups despite gazing at a single target (e.g., looking at a partner in a side-conversation while remaining part of a broader TV-watching group).

Finally, our method integrates individual attention heatmap $\bm{A}$ and $\bm{M}$ to yield a SA heatmap for each group:
\begin{align}
    \bm{S}_{e} = \frac{1}{N} \sum_{n} (\bm{M}_{e,n} \cdot \bm{A}_{n})
    \label{eq:init_sae}
\end{align}
where $\bm{S}_{e}$ and $\bm{M}_{e,n}$ denote the $e$-th group's SA heatmap and the $n$-th person's group membership, respectively.

\subsection{Refined Shared Attention Heatmap Estimation}
\label{subsec:prop_refine_sae}

\paragraph{Key idea.}

Obtaining $\bm{S}_{e}$ via a one-way process (i.e., group detection then SA estimation) is insufficient, as group detection lacks access to the subsequent SA estimation.
Thus, individual attentions can be inconsistent within the detected group.
To resolve the issue, we propose refinement of group detection with feedback from $\bm{S}$, as shown in Fig.~\ref{fig:sae_refinement}.

\vspace{-3mm}

\paragraph{Details.}
Specifically, spatial argmax is applied to $\bm{S}_{e}$ to obtain coordinates indicating the maximum values (i.e., peaks), as these coordinates efficiently represent information of the initial shared attention estimation.
These coordinates are concatenated to $\bm{U}$ to obtain refined group tokens $\bm{U}' \in \mathbb{R}^{E \times (D+2)}$.
$\bm{U}'$ is fed into the group detector to obtain refined memberships $\bm{M}' \in \mathbb{R}^{E \times N}$ as with Eq.~\ref{eq:groupness}:
\begin{align}
    \bm{M}'_{e,n} = \mathrm{Sigmoid} (\mathrm{g'}(U_{e}') \cdot \mathrm{f'}(P'_{n}))
    \label{eq:refined_group_prob}
\end{align}
where $\mathrm{g'}$ and $\mathrm{f'}$ denote FC layers to embed $\bm{U'}$ and $\bm{P'}$, respectively.
By using $\bm{M'}$, refined shared attention heatmap $\bm{S}' \in \mathbb{R}^{E \times H \times W}$ is recomputed as with Eq.~\ref{eq:init_sae}:
\begin{align}
    \bm{S}'_{e} = \frac{1}{N} \sum_{n} (\bm{M}'_{e,n} \cdot \bm{A}_{n})
    \label{eq:refined_sae}
\end{align}

During inference, the $e$-th group is identified as an actual SA group if multiple $\bm{M}'_{e,n}$ exceed a predefined threshold among $N$ people (i.e., $|\{n \mid \bm{M}'_{e,n} > \tau\}| \geq 2$, where $\tau$ is a predefined threshold). Otherwise, the group is discarded in the subsequent shared attention estimation process.
During training, instead of thresholding, we ensure end-to-end differentiability by integrating $\bm{A}$ with $\bm{M}$ to estimate $\bm{S}$, as described in Sec~\ref{subsec:prop_init_sae}.

\subsection{Loss functions}
\label{subsec:prop_losses}

The network is trained on the following loss function $\mathcal{L}$:
\begin{align}
    \mathcal{L}_{grp} &= \mathrm{BCE}(\bm{M}', \bm{M}_{gt}) + \mathrm{BCE}(\bm{M}, \bm{M}_{gt}) \\ 
    \mathcal{L}_{sae} &= \mathrm{MSE}(\bm{S}', \bm{S}_{gt}) + \mathrm{MSE}(\bm{S}, \bm{S}_{gt}) \\
    \mathcal{L} &= \mathcal{L}_{grp} + \mathcal{L}_{sae} + \mathcal{L}_{aux}
    \label{eq:losses}
\end{align}
where $\mathcal{L}_{grp}$ and $\mathcal{L}_{sae}$ are group detection and shared attention estimation losses, respectively. 
$\mathrm{BCE}$ and $\mathrm{MSE}$ are binary cross entropy and mean-square error.
Following~\cite{DBLP:conf/eccv/CarionMSUKZ20,DBLP:conf/eccv/KimSCK24}, the Hungarian algorithm~\cite{hungarian} performs bipartite matching between detected and ground-truth groups.
This optimization sorts $\bm{M}'$, $\bm{S}'$, $\bm{M}_{gt}$, and $\bm{S}_{gt}$ to minimize $\mathcal{L}_{grp}$.

The auxiliary loss $\mathcal{L}_{aux}$ consists of $\mathcal{L}_{ang}$, $\mathcal{L}_{hm}$, $\mathcal{L}_{io}$, and $\mathcal{L}_{social}$, all of which are used in~\cite{DBLP:conf/nips/GuptaTFVO24}.
$\mathcal{L}_{ang}$ and $\mathcal{L}_{hm}$ supervise individual gaze direction and  heatmap, $\mathcal{L}_{io}$ accounts for gaze existence within the frame, and $\mathcal{L}_{social}$ is designed to capture social interactions between individuals.
\section{Experimental results}
\label{sec:exp}

\subsection{Dataset}
\label{subsec:exp_dataset}

We rely on the VSGaze dataset~\cite{DBLP:conf/nips/GuptaTFVO24} which is composed of four sub-datasets: VAT~\cite{DBLP:conf/cvpr/ChongWRR20}, ChildPlay~\cite{DBLP:conf/iccv/TafascaGO23}, VideoCoAtt~\cite{DBLP:conf/cvpr/FanCWWZ18}, and UCO-LAEO~\cite{DBLP:conf/cvpr/Marin-JimenezKM19}.
In all datasets, the head bounding boxes of each person and their 2D gaze target are annotated.
Since UCO-LAEO contains no shared attention labels, we only exploit the three other sub-datasets (i.e., VideoCoAtt, VideoAttentionTarget, and ChildPlay), by exploiting their shared attention annotation labels and their group members.

\noindent{\bf VideoCoAtt~\cite{DBLP:conf/cvpr/FanCWWZ18}} was originally proposed for shared attention estimation. It consists of 380 videos from TV shows, capturing natural social interactions. 
%
In these videos, frames containing shared attention are annotated with corresponding bounding boxes.

\noindent{\bf VideoAttentionTarget~\cite{DBLP:conf/cvpr/ChongWRR20}} contains 1331 clips collected from 50 shows on YouTube, including movie clips, sitcoms, and TV shows. Each clip is extracted from 50 source videos and ranges from 1 to 80 seconds. 
Unlike VideoCoAtt, which includes the bounding box annotations of shared attention, Gupta~\cite{DBLP:conf/nips/GuptaTFVO24} automatically annotates shared attention by checking whether each person's 2D gaze targets fall within the same head box.

\noindent{\bf ChildPlay~\cite{DBLP:conf/iccv/TafascaGO23}} collects 95 YouTube videos with queries (e.g., ``children playing toys'') and obtains 401 clips. In each clip, at least 1 child is interacting with other children and adults. As with VideoAttentionTarget, shared attention bounding box is automatically annotated from each person's 2D gaze targets.

VideoCoAtt (400k/918k), VideoAttentionTarget (16k/94k), and ChildPlay (4k/55k) provide the positive and negative samples, respectively.
While a positive sample indicates that the image contains at least one shared attention, a negative sample means that the image includes no shared attention.
As VideoCoAtt is the predominant source of the VSGaze dataset, we report results on each dataset individually for detailed analysis.

By using the annotations for group detection and shared attention, including the shared attention bounding box and group of people focusing on it, we prepare the ground-truth group membership and shared attention heatmaps (i.e., $\bm{M}_{gt}$ and $\bm{S}_{gt}$) required for network optimization.

\begin{table*}[t]
    \centering
    \begin{tabular}{c|c|c|c|c|c|c} \hline
        & \multicolumn{6}{c}{GroupAP} \\ \hline
        & \multicolumn{3}{c|}{$\theta_{\mathrm{IoU}}$=0.5} & \multicolumn{3}{c}{$\theta_{\mathrm{IoU}}=1.0$} \\ \hline
        Method & $\theta_{\mathrm{Dist}}=0.05$ & $\theta_{\mathrm{Dist}}=0.1$ & $\theta_{\mathrm{Dist}}=\infty$ & $\theta_{\mathrm{Dist}}=0.05$ & $\theta_{\mathrm{Dist}}=0.1$ & $\theta_{\mathrm{Dist}}=\infty$ \\ \hline
        MTGS-PP~\cite{DBLP:conf/nips/GuptaTFVO24} & 16.4 & 20.7 & 37.5 & 7.1 & 8.6 & 11.3 \\ \hline
        MTGS-Soc.~\cite{DBLP:conf/nips/GuptaTFVO24} & 5.7 & 7.8 & 28.6 & 2.9 & 3.6 & 9.8 \\ \hline
        Gaze-LLE-PP~\cite{DBLP:conf/cvpr/RyanB0BHR25} & 15.6 & 19.5 & 26.3 & 5.7 & 6.9 & 8.2 \\ \hline
        Ours & \red{32.4} & \red{41.0} & \red{61.7} & \red{12.5} & \red{15.9} & \red{17.7} \\ \hline
    \end{tabular}
    \caption{Comparison of previous and our methods on the VideoCoAtt dataset. The best result in each column is colored by \red{red}. Regarding the group detection, $\theta_{\mathrm{IoU}}=0.5$ and $\theta_{\mathrm{IoU}}=1.0$ are used for lenient and strict criteria.
    For the shared attention estimation, $\theta_{\mathrm{Dist}}=0.05$, $\theta_{\mathrm{Dist}}=0.1$, and $\theta_{\mathrm{Dist}}=\infty$ are used for evaluation. $\theta_{\mathrm{Dist}}=0.05$ and $\theta_{\mathrm{Dist}}=0.1$ judge whether estimated shared attention is located within a pre-defined distance threshold from ground-truth shared attentions. $\theta_{\mathrm{Dist}}=\infty$ is used to only evaluate group detection while ignoring shared attention estimation errors. 
    }
    \label{table:exp_comp_coatt}
\end{table*}

\begin{figure*}
    \centering
    \includegraphics[width=\textwidth]{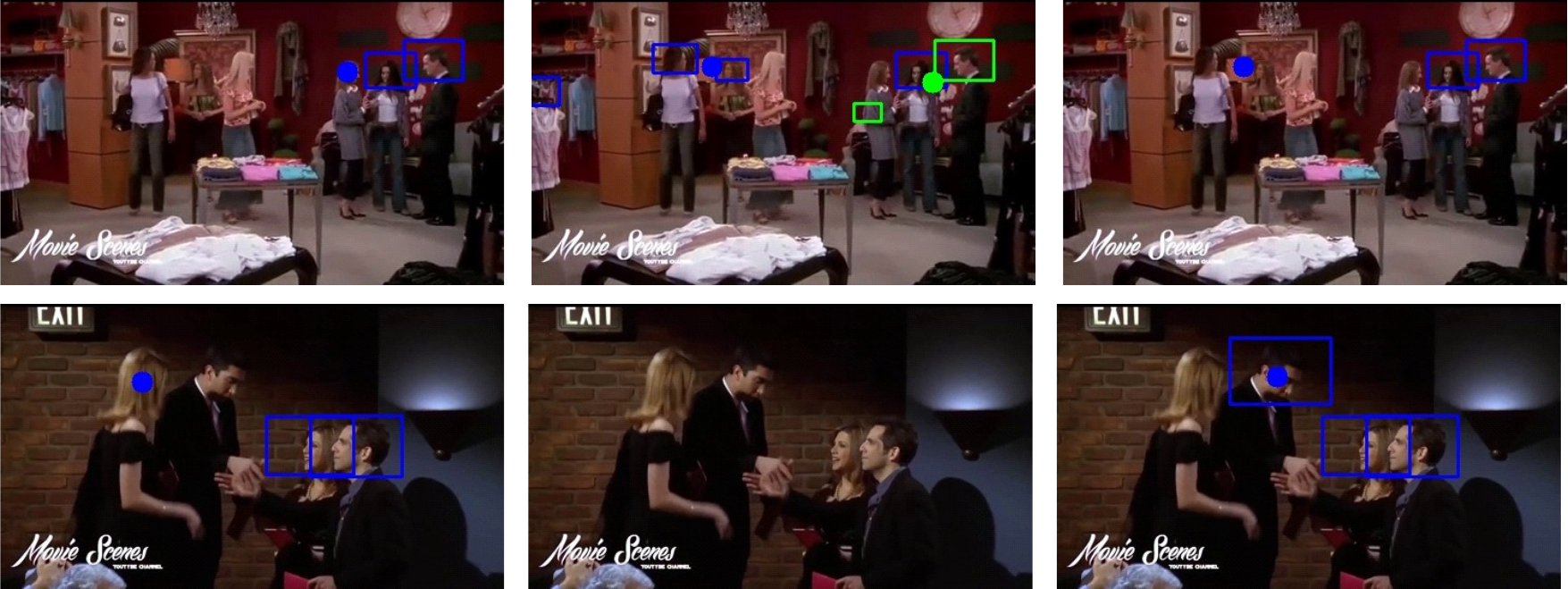}
    Ground-truth~\hspace{40mm}
    MTGS-PP~\hspace{40mm}
    Ours
    \caption{Visual comparison on the VideoCoAtt dataset.
    People joining the same group are enclosed within a rectangle of the same color. The estimated shared attention point (i.e., the peak of the estimated shared attention heatmap) for each group is visualized as each group's color point.
    Due to space limitations, we compare our method with MTGS-PP, having the second-best performance in Table~\ref{table:exp_comp_coatt}.
    }
    \label{fig:exp_vis_comp_coatt}
\end{figure*}

\begin{table*}[t]
    \centering
    \begin{tabular}{c|c|c|c|c|c|c} \hline
        & \multicolumn{6}{c}{GroupAP} \\ \hline
        & \multicolumn{3}{c|}{$\theta_{\mathrm{IoU}}$=0.5} & \multicolumn{3}{c}{$\theta_{\mathrm{IoU}}=1.0$} \\ \hline
        Method & $\theta_{\mathrm{Dist}}=0.05$ & $\theta_{\mathrm{Dist}}=0.1$ & $\theta_{\mathrm{Dist}}=\infty$ & $\theta_{\mathrm{Dist}}=0.05$ & $\theta_{\mathrm{Dist}}=0.1$ & $\theta_{\mathrm{Dist}}=\infty$ \\ \hline
        MTGS-PP~\cite{DBLP:conf/nips/GuptaTFVO24} & \red{30.3} & \red{35.6} & \red{66.0} & 5.1 & 6.3 & 9.0 \\ \hline
        MTGS-Soc.~\cite{DBLP:conf/nips/GuptaTFVO24} & 7.6 & 9.9 & 37.2 & 1.5 & 1.9 & 4.7 \\ \hline
        Gaze-LLE-PP~\cite{DBLP:conf/cvpr/RyanB0BHR25} & 7.5 & 8.5 & 12.4 & 2.9 & 3.0 & 3.3 \\ \hline
        Ours & 27.2 & 33.5 & 52.0 & \red{7.5} & \red{10.1} & \red{12.0} \\ \hline
    \end{tabular}
    \caption{Comparison of previous and our methods on the VideoAttentionTarget dataset. The best result in each column is colored by \red{red}.}
    \label{table:exp_comp_vat}
\end{table*}

\begin{figure*}
    \centering
    \includegraphics[width=\textwidth]{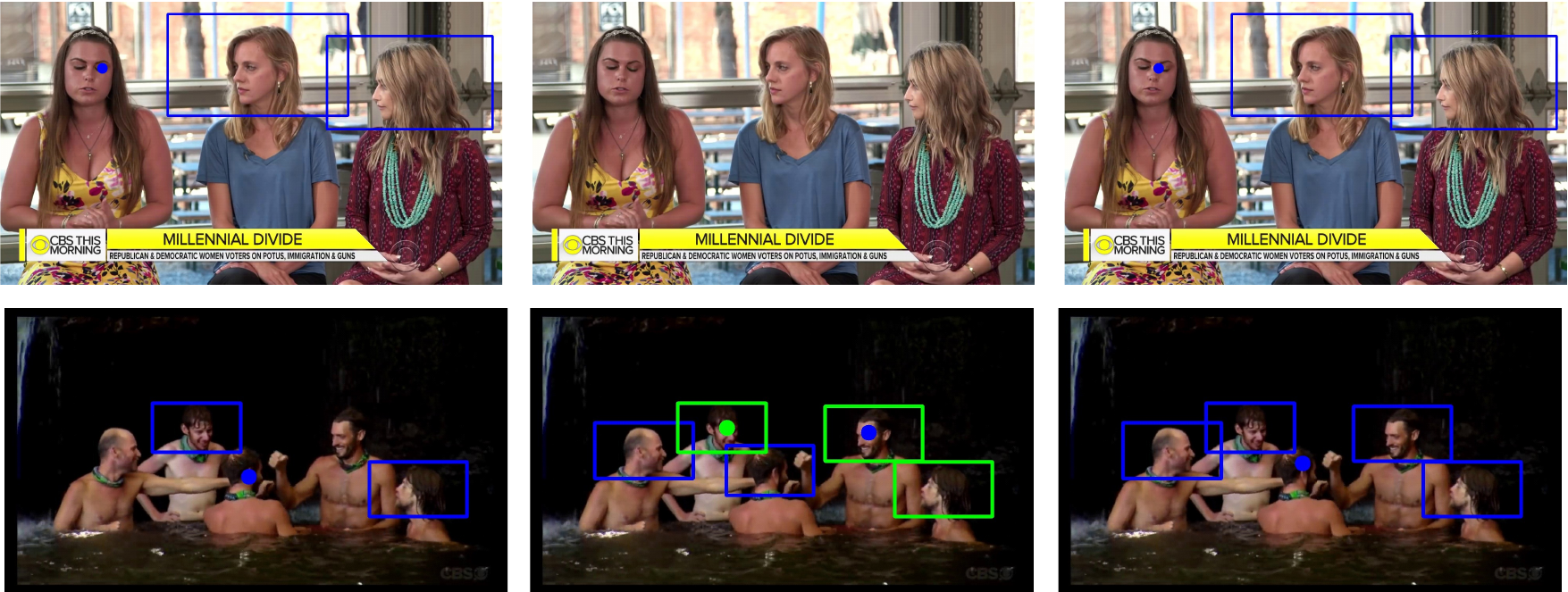}
    Ground-truth~\hspace{40mm}
    MTGS-PP~\hspace{40mm}
    Ours
    \caption{Visual comparison on the VideoAttentionTarget dataset. 
    }
    \label{fig:exp_vis_comp_vat}
\end{figure*}

\begin{table*}[t]
    \centering
    \begin{tabular}{c|c|c|c|c|c|c} \hline
        & \multicolumn{6}{c}{GroupAP} \\ \hline
        & \multicolumn{3}{c|}{$\theta_{\mathrm{IoU}}$=0.5} & \multicolumn{3}{c}{$\theta_{\mathrm{IoU}}=1.0$} \\ \hline
        Method & $\theta_{\mathrm{Dist}}=0.05$ & $\theta_{\mathrm{Dist}}=0.1$ & $\theta_{\mathrm{Dist}}=\infty$ & $\theta_{\mathrm{Dist}}=0.05$ & $\theta_{\mathrm{Dist}}=0.1$ & $\theta_{\mathrm{Dist}}=\infty$ \\ \hline
        MTGS-PP~\cite{DBLP:conf/nips/GuptaTFVO24} & 7.8 & 13.9 & 25.4 & \red{4.6} & \red{5.2} & \red{5.6} \\ \hline
        MTGS-Soc.~\cite{DBLP:conf/nips/GuptaTFVO24} & 0.8 & 1.8 & 9.1 & 0.7 & 0.7 & 1.3 \\ \hline
        Gaze-LLE-PP~\cite{DBLP:conf/cvpr/RyanB0BHR25} & 5.8 & 8.1 & 14.1 & 2.4 & 2.8 & 4.4 \\ \hline
        Ours & \red{9.0} & \red{15.6} & \red{36.3} & 2.1 & 2.1 & 2.4 \\ \hline
    \end{tabular}
    \caption{Comparison of previous and our methods on the ChildPlay dataset. The best result in each column is colored by \red{red}.}
    \label{table:exp_comp_child}
\end{table*}

\begin{figure}
    \centering
    \includegraphics[width=\columnwidth]{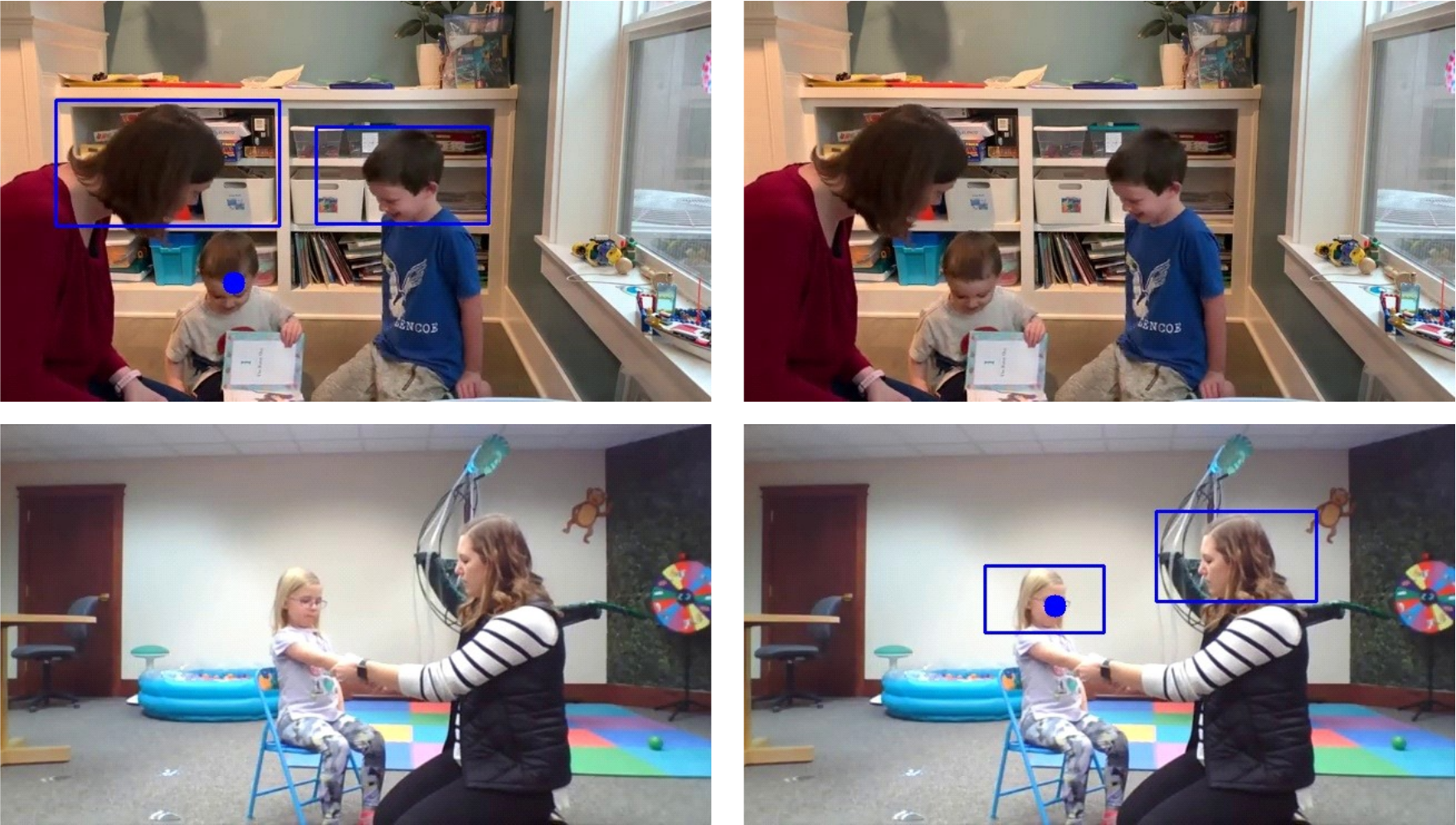}
    Ground-truth~\hspace{30mm}
    Ours
    \caption{Failure cases on the ChildPlay dataset.}
    \label{fig:exp_vis_failure_child}
\end{figure}

\subsection{Evaluation metrics}
\label{subsec:exp_eval}

Following~\cite{DBLP:conf/eccv/KimSCK24}, we evaluate using GroupAP, but introduce dual criteria to specialize the metric for simultaneous group detection and shared attention estimation tasks.

Specifically, each detection is regarded as successful if GroupIoU is larger than $\theta_{\mathrm{IoU}}$ and the GroupDist is less than $\theta_{\mathrm{Dist}}$.
GroupIoU and GroupDist in the dual criteria evaluate the group detection and shared attention estimation, respectively.
$\theta_{\mathrm{IoU}}$ and $\theta_{\mathrm{Dist}}$ denote the thresholds for group detection and shared attention estimation.
The thresholds determine whether each task is regarded as a success or a failure.
GroupIoU and GroupDist are computed as follows:
\begin{align}
    \mathrm{GroupIoU}(\bm{M}', \bm{M}_{gt}) &= \frac{|\bm{M}' \cap \bm{M}_{gt}|}{|\bm{M}' \cup \bm{M}_{gt}|} \\
    \mathrm{GroupDist}(\bm{S}', \bm{p}_{gt}) &= \| \bm{p}' - \bm{p}_{gt} \|_2
    \label{eq:exp_eval_metricx}
\end{align}
Here, $\bm{p}'$ denotes the $(x,y)$ coordinates corresponding to the maximum value in $\bm{S}'$, and $\bm{p}_{gt}$ denotes the $(x,y)$ coordinates of the ground-truth shared attention. Both coordinates are normalized by the image width $W$ and height $H$ (i.e., $(x/W, y/H)$).
AP is computed using the dual criteria. Here, the maximum value of $\bm{S'}$ serves as the confidence score for each joint estimation of shared attention and group detection obtained from a single group token.

In Sec.~\ref{subsec:exp_comp}, we report results with different thresholds: $\theta_{\mathrm{IoU}} \in \{0.5, 1.0\}$ for group detection and $\theta_{\mathrm{Dist}} \in \{0.05, 0.1, \infty\}$ pixels for shared attention estimation. 
The $\theta_{\mathrm{Dist}}=\infty$ setting ignores shared attention estimation and validates only group detection performance.

\subsection{Experimental details}
\label{subsec:exp_details}

Following Gupta~\textit{et al.}, our network is trained in two stages. 
In the first stage, we only train the individual attention estimation (Sec.~\ref{subsec:prop_single_att}) on the GazeFollow dataset~\cite{DBLP:conf/nips/RecasensKVT15}.
In the second stage, we train the whole network on the VSGaze dataset.
Regarding other experimental settings, we follow Gupta~\textit{et al.}.
For model optimization, we utilize the Adam~\cite{DBLP:journals/corr/Adam} with a learning rate of $1 \times 10^{-5}$ in both stages. 

\subsection{Comparative experiments}
\label{subsec:exp_comp}

\subsubsection{Compared methods}
\label{subsec:exp_comp_methods}

Since no existing methods simultaneously estimate shared attention and group members, we extend MTGS~\cite{DBLP:conf/nips/GuptaTFVO24} and GAZE-LLE~\cite{DBLP:conf/cvpr/RyanB0BHR25} for comparison, as these individual attention estimation methods are more adaptable than direct shared attention estimation methods~\cite{DBLP:conf/cvpr/FanCWWZ18,DBLP:journals/tcsv/ZhuangNXYZLG20,DBLP:conf/iccv/NakataniKU23,DBLP:conf/wacv/SumerGTK20}.

\noindent{\bf MTGS-Soc.} 
MTGS only estimates pairwise shared attention labels (e.g., $1$st and $2$nd people share their attention). 
To extend this pairwise estimation to group detection for a fair comparison, we use graph clustering as in~\cite{DBLP:conf/iccv/YokoyamaKU25}. Using the Louvain algorithm~\cite{DBLP:journals/corr/abs-2311-06047} as graph clustering, we obtain group detection results from pairwise estimation.

\noindent{\bf MTGS-PP.} 
We include an MTGS variant with post-processing to derive SA groups. If the estimated attention points of two or more people fall within a distance threshold empirically set for each sub-dataset, we regard them as sharing attention and assign them to the same group.

\noindent{\bf Gaze-LLE-PP.}
As with MTGS-PP, we also apply the same post-processing strategy to Gaze-LLE and use it as another comparison baseline.

\subsubsection{Results on VideoCoAtt}
\label{subsubsec:exp_comp_coatt}

Table~\ref{table:exp_comp_coatt} shows the comparison of previous and our methods on the VideoCoAtt dataset.
These results validate that our method outperforms previous methods across all settings on the VideoCoAtt dataset.
Our method achieves $32.4\%$ GroupAP ($\theta_{\mathrm{IoU}}=0.5$, $\theta_{\mathrm{Dist}}=0.05$) versus $16.4\%$ for MTGS-PP.
Furthermore, results at $\theta_{\mathrm{Dist}}=\infty$ ($61.7\%$ vs $37.5\%$) confirm that our method achieves high performance solely through group detection.

Under strict settings ($\theta_{\mathrm{IoU}}=1.0, \theta_{\mathrm{Dist}}=0.05$), we achieve $12.5\%$ versus $7.1\%$ for MTGS-PP.
These results indicate that our method is best in both lenient (i.e., $\theta_{\mathrm{IoU}}=0.5$) and strict (i.e., $\theta_{\mathrm{IoU}}=1.0$) settings.
The small improvement between results at $\theta_{\mathrm{Dist}}=0.05$ and $\theta_{\mathrm{Dist}}=\infty$ indicates that detecting groups perfectly is difficult. This is because the ground-truth annotations themselves are ambiguous and difficult to learn group detection.

Figure~\ref{fig:exp_vis_comp_coatt} shows visual comparisons between MTGS-PP and our method on the VideoCoAtt dataset.
In this figure, people detected as the same group are enclosed by the same color rectangle, and their estimated shared attention is visualized by a point with the group's color. 
 
In the upper example, our method perfectly detects the ground-truth group, whereas MTGS-PP misdetects groups. However, the estimated shared attention, visualized as a blue point, is far from the ground-truth point. The result indicates the difficulty of this task; the shared attention estimation sometimes fails even with perfect group detection.

In the bottom example, while MTGS-PP detects no groups, our method correctly identifies a group to some extent (i.e., GroupIoU is improved from 0.00 to 0.66). 

While detecting the correct group and localizing shared attention are difficult due to the complexity of this task, our method outperforms other methods, as validated in Table~\ref{table:exp_comp_coatt} and Fig.~\ref{fig:exp_vis_comp_coatt} on the VideoCoAtt dataset.

\subsubsection{Results on VideoAttentionTarget}
\label{subsubsec:exp_comp_vat}

Table~\ref{table:exp_comp_vat} shows that our method excels under strict criteria on the VideoAttentionTarget dataset.
Specifically, our method achieves $7.5\%$ at $\theta_{\mathrm{IoU}}=1.0$ versus $5.1\%$ for MTGS-PP (47\% improvement).
Under lenient settings ($30.3\%$ vs $27.2\%$ at $\theta_{\mathrm{IoU}}=0.5$), while MTGS-PP achieves higher GroupAP, our method still maintains second-best results. 

The strict criteria ($\theta_{\mathrm{IoU}}=1.0$) provide a more rigorous assessment of group detection than the lenient criteria ($\theta_{\mathrm{IoU}}=0.5$). Since our method outperforms MTGS-PP under strict settings, this demonstrates that it produces more accurate group detection, despite a minor trade-off in lenient criteria where precise group detection is not required.

In the upper example of Fig.~\ref{fig:exp_vis_comp_vat},  MTGS-PP fails to detect the shared-attention group, whereas our method correctly detects it and estimates the shared-attention point.
In the bottom example, MTGS-PP splits two people who share attention into different groups. 
Our method still fails to completely detect the ground-truth group, but the two people are correctly detected as being part of the same group.

\subsubsection{Results on ChildPlay}
\label{subsubsec:exp_comp_child}

Table~\ref{table:exp_comp_child} shows our method achieves best performance under lenient criteria ($36.3\%$ at $\theta_{\mathrm{Dist}}=\infty$ vs MTGS-PP's $25.4\%$). 
The low performance even at $\theta_{\mathrm{Dist}}=\infty$ comes from the dataset composition. The ChildPlay dataset contains a significantly higher proportion of negative (non-shared attention) samples compared to other datasets, as described in Sec.~\ref{subsec:exp_dataset}. 
In the negative-dominated dataset, identifying the rare actual groups is fundamentally difficult.

Concretely, the ChildPlay dataset has 4k/55k positive/negative samples, indicating only 7.3\% positive samples, whereas VideoCoAtt has 43.5\% positive samples, and VideoAttentionTarget has 17.0\% positive samples. 
This class imbalance makes group detection inherently challenging, limiting overall performance even when strict attention estimation is not required.

Failure cases on the ChildPlay dataset are shown in Fig.~\ref{fig:exp_vis_failure_child}.
In the upper example, our method fails to detect any groups, while the ground-truth includes one group. This failure is likely due to the significant class imbalance in the ChildPlay dataset.

In the bottom example, our method incorrectly estimates the shared attention point above the person in the detected group, while there is no shared attention in the ground-truth.
This is problematic because the shared attention point should not be located on a person within the detected group, suggesting that our method may not have fully learned the fundamental nature of shared attention estimation.

\begin{table}[t]
    \centering
    \begin{tabular}{c|c|c|c} \hline
        & \multicolumn{3}{c}{GroupAP} \\ \hline
        & \multicolumn{3}{c}{$\theta_{\mathrm{IoU}}$=1.0} \\ \hline
        Method & $\theta_{\mathrm{Dist}}=0.05$ & $\theta_{\mathrm{Dist}}=0.1$ & $\theta_{\mathrm{Dist}}=\infty$ \\ \hline\hline
        w/o refinement & 2.5 & 3.3 & 3.6 \\ \hline
        w/o $\mathcal{L}_{social}$ & 0.9 & 1.6 & 1.9 \\ \hline\hline
        Ours (Full) & \red{12.5} & \red{15.9} & \red{17.7} \\ \hline
    \end{tabular}
    \caption{Ablation study on the VideoCoAtt dataset. 
    }
    \label{table:exp_abl}
\end{table}

\subsection{Ablation study}
\label{subsec:exp_abl}

We conduct the following ablation studies on the VideoCoAtt dataset to assess each component in our method.

\paragraph{Effectiveness of refinement (Sec.~\ref{subsec:prop_refine_sae})}
We ablate the shared attention refinement (i.e., Sec.~\ref{subsec:prop_refine_sae}), where the initial shared attention $\bm{S}$ is directly regarded as final shared attention estimations.
Table~\ref{table:exp_abl} shows that our refinement of group detection and shared attention estimation certainly improves the GroupAP in all $\theta_{\mathrm{Dist}}$ settings.

\begin{figure}
    \centering
    \includegraphics[width=\columnwidth]{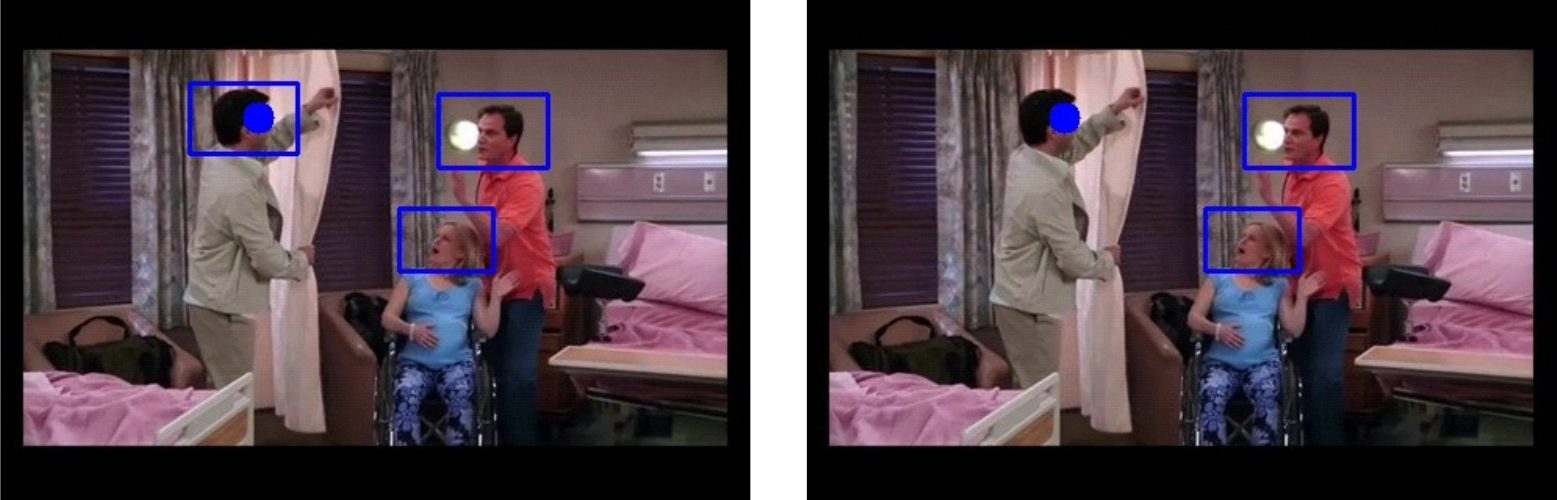}
    Initial estimation ($S$)~\hspace{10mm}
    Refined estimation ($S'$)
    \caption{Effectiveness of refined shared attention estimation (Sec.~\ref{subsec:prop_refine_sae}) on the VideoCoAtt dataset.}
    \label{fig:exp_vis_refine}
\end{figure}

Figure~\ref{fig:exp_vis_refine} also validates the effectiveness of refinement. 
In the initial estimation ($S$), the leftmost person is incorrectly detected within the group, even though the shared attention is estimated above the leftmost person. This is because initial group detection cannot access the final estimation of shared attention. However, in the refined estimation ($S'$), the leftmost person is removed to refine group detection using feedback from the initial shared attention estimation.

\paragraph{Effectiveness of $\mathcal{L}_{social}$}
We analyze the effectiveness of $\mathcal{L}_{\mathrm{social}}$ in $\mathcal{L}_{aux}$. Since Table~\ref{table:exp_abl} shows that $\mathcal{L}_{\mathrm{social}}$ enhances the GroupAP in all settings, we can say that learning pairwise social interactions between people is also beneficial for shared attention estimation, even though the main task is group detection and shared attention estimation.

\section{Concluding Remarks}
\label{sec:conclusion}

This paper proposes a new approach for shared attention estimation.
Previous shared attention estimators neither fail to detect the actual SA group members nor assume multiple SA points.
Unlike these estimators, we propose to estimate shared attentions via group detection, enabling the estimation of multiple SA points along with their actual group members.
Experimental results on the VSGaze dataset demonstrate the superiority of our method.

Future work includes refinements of individual attention by leveraging group context and shared attention results.
Future work includes large-scale data collection and cross-dataset evaluations to further validate the robustness and generalization of our method in diverse environments.

{
    \small
    \bibliographystyle{ieeenat_fullname}
    \bibliography{main}
}


\end{document}